\makeatletter
\disable@package@load{program}{}
\makeatother

\documentclass[iicol,pdflatex,sn-mathphys-num]{sn-jnl}

\usepackage{graphicx}%
\usepackage{multirow}%
\usepackage{amsmath,amssymb,amsfonts}%
\usepackage{amsthm}%
\usepackage{mathrsfs}%
\usepackage[title]{appendix}%
\usepackage{xcolor}%
\usepackage{textcomp}%
\usepackage{manyfoot}%
\usepackage{booktabs}%
\usepackage{algorithm}%
\usepackage{algorithmicx}%
\usepackage{algpseudocode}%
\usepackage{listings}%
\usepackage{comment}

\usepackage{url} 
\usepackage{pifont} 
\usepackage{bm}
\usepackage{array}

\usepackage{hyperref}

\let\orcidlogo\relax
\usepackage{orcidlink}
\renewcommand{\orcid}[1]{}

\theoremstyle{thmstyleone}%

\theoremstyle{thmstyletwo}%

\theoremstyle{thmstylethree}%

\usepackage{enumitem}
\setlist{nosep, leftmargin=*} 

\renewcommand{\arraystretch}{0.95} 
\usepackage{listings}
\lstdefinestyle{code}{
  basicstyle=\ttfamily\footnotesize,
  breaklines=true,
  columns=fullflexible,
  keepspaces=true,
  showstringspaces=false,
  frame=single,
  framesep=3pt,
  xleftmargin=4pt,
  xrightmargin=4pt,
}
\lstdefinestyle{code}{
    columns=fixed,
    basewidth={0.45em,0.42em},  
    keepspaces=true,
    frame=single,          
    framesep=0pt,          
    rulesep=0pt,           
    xleftmargin=0pt,       
    xrightmargin=0pt,      
    rulecolor=\color{gray},
}

\usepackage{seqsplit}
\usepackage{url}

\begin{document}

\title[]{Grounding LLM Agents in Business Processes: Process-Aware Knowledge Graphs for Reliable Multi-Step Service Invocation}

\author*[1]{\fnm{Saiful} \sur{Khan}\orcidlink{0000-0002-6796-5670}}\email{saiful.khan@stfc.ac.uk}
\author[2]{\fnm{Afrah} \sur{Farea}\orcidlink{0000-0003-4412-5377}}

\affil*[1]{\orgdiv{Scientific Computing}, \orgname{Science and Technology Facilities Council}, \orgaddress{\street{RAL}, \postcode{OX11 0QX}, \country{UK}}}

\affil[2]{\orgdiv{Department of Engineering Science}, \orgname{University of Oxford}, \orgaddress{\street{Parks Road}, \postcode{OX1 3PJ}, \country{UK}}}


\abstract{
Large language model (LLM) agents that call enterprise APIs must select the correct endpoints, follow process order, use valid payload keys, and choose coded values from predefined lists. However, current retrieval-augmented service discovery treats endpoints independently and therefore misses process and schema constraints. We address this limitation by integrating Open Data Protocol (OData) metadata and Business Process Model and Notation (BPMN) reference processes in a typed property graph. A hybrid retriever identifies candidate endpoints using dense embeddings and follows process edges to retrieve subsequent endpoints.
We evaluate the method on 42 enterprise services with process-chain ground truth. For queries with low lexical overlap between the query and downstream entities, it improves Recall@3 by 5.7 points ($n=74$), with consistent gains across embedding models and cut-offs. It matches dense retrieval on other queries and provides no improvement on OpenAPI services without a process layer, indicating that the gains arise from process structure.
We also evaluate an LLM agent on 81 multi-step queries. It completes 38.7\% of tasks, compared with 10.3\% for a chunk-based baseline. In an elicitation probe, providing valid code lists eliminates invented codes but leaves invalid keys in 46\% of write runs. Full typed context reduces invalid calls to 3\% (3 of 101 runs), with no invented codes. These results show that exposing process and schema constraints during retrieval improves the reliability of agent-driven API calls.
}



\maketitle

\section{Introduction}
\label{sec:introduction}

Imagine an AI assistant that must arrange a business trip. It has to find a flight, book it, and then reserve a hotel using the booking reference, in that order. Each booking form also has required fields and fixed code lists, such as cabin class or fare type. The assistant can pick the right booking API and still fail: it may invent a fare code, leave out a required reference, or book the hotel before the flight. Such a call fails just as badly as a call to the wrong API. Reliable multi-step automation therefore needs more than the right endpoint. It also needs two kinds of constraints around each call: \emph{sequential} constraints (what comes next) and \emph{parametric} constraints (which keys and code values are valid).

Enterprise software has the same problem at a much larger scale. Large companies expose thousands of APIs. A typical SAP S/4HANA system, for example, ships hundreds of Open Data Protocol (OData) services, each with many collections, fields, and cross-references. Suppose a developer asks: \emph{``create a purchase order for supplier X with a valid status code, then confirm goods receipt.''} Choosing the right endpoint is only the first step. The payload keys must match the schema, the coded values must be valid, and the calls must follow the business process. Each of these rules gives a language model a chance to produce a call that looks right but is invalid.

Recent work on retrieval-augmented service discovery~\cite{Pesl:IEEE-TSC:2026,song2023restgpt} improves endpoint retrieval with chunking and embedding-based filtering. However, it treats each endpoint as an isolated unit. It leaves out three kinds of context: (a) \textbf{process}, what comes next in the workflow; (b) \textbf{structure}, which entities and value-help codes are valid; and (c) \textbf{execution}, how an agent builds and sends the call. The Discovery Agent~\cite{Pesl:IEEE-TSC:2026}, for example, ranks endpoints but stops at retrieval, and it has no process-chain ground truth. Its authors suggest knowledge graphs as a way to improve accuracy, but note that graphs are costly to build. We address this gap.

Our approach links two open, machine-readable standards. OData \texttt{\$metadata} describes the entity collections, fields, and references of a service, but not the order in which they are used. BPMN describes the order of business tasks, but not the data each task uses. We load both sources into one typed property graph and link process activities to the entity sets they use. A hybrid retriever finds the closest entity sets by embedding similarity and then follows process edges to the ones that come next. It returns each endpoint with its schema context, which an execution agent uses to issue validated GET and POST calls through a mock HTTP layer.

This paper makes four contributions:

\begin{enumerate}
\item[(a)]
A typed property-graph schema that combines service metadata with business processes, and an ingestion pipeline for OData \texttt{\$metadata} and BPMN 2.0. We evaluate this design with a construction-fidelity audit (Sec.~\ref{sec:eval-fidelity}) and an ablation study (Sec.~\ref{sec:eval-rq2}).

\item[(b)] A hybrid retriever that combines dense embedding seeds with a bounded expansion along process-sequence edges. We evaluate it in Sec.~\ref{sec:eval-rq1} and test where it stops working in Sec.~\ref{sec:eval-rq4}.

\item[(c)] An execution-capable ReAct~\cite{yao:2023:ICLR:react} agent that uses the retrieved value-help and cardinality context to build its calls. We evaluate it in Secs.~\ref{sec:eval-rq3} and~\ref{sec:eval-rq3b}, including a baseline that receives the value lists as plain text, which separates the effect of seeing the values from the effect of the typed representation.

\item[(d)] \emph{OData-Proc}, a public benchmark of $42$ enterprise services, $14$ BPMN reference processes, and $288$ queries with two-annotator, process-chain ground truth. To our knowledge it is the first benchmark that scores service discovery against business-process chains.
\end{enumerate}

Sec.~\ref{sec:background} explains the main terms for readers who are new to OData, or BPMN. Sec.~\ref{sec:relwork} reviews related work, Sec.~\ref{sec:approach} describes our method, and Sec.~\ref{sec:eval} reports the experiments.

\section{Background}
\label{sec:background}

This section introduces the systems and standards we build on; Table~\ref{tab:glossary} summarises the key terms and metrics.

\begin{table}[!htb]
\centering
\caption{Key terms and metrics used in this paper.}
\label{tab:glossary}
\footnotesize
\renewcommand{\arraystretch}{1.1}
\begin{tabular}{@{}p{1.7cm}p{5.45cm}@{}}
\toprule
\textbf{Term} & \textbf{Meaning} \\
\midrule
Service (API) & A web interface that lets a program read or write business data. \\
\texttt{\$metadata} & The machine-readable description that every OData service publishes. \\
EntitySet & A named collection of records, similar to a table (e.g., \texttt{PurchaseOrder}). \\
Property & One field of a record (e.g., \texttt{Supplier}). \\
Navigation & A link between two EntitySets, with a cardinality such as 1:N. \\
Value help & The fixed list of valid codes for a property (e.g., \texttt{NB}, \texttt{UB}, \texttt{FO}). \\
Payload & The property values sent in a POST call that creates a record. \\
Activity & One task in a BPMN process (e.g., \emph{create purchase order}). \\
Mapping edge & A hand-written link from an activity to an EntitySet it reads or writes. \\
Process chain & The EntitySets used by an activity and by the activities that follow it. \\
Invalid call & A GET or POST call that uses an EntitySet, key, or code that does not exist. \\
\midrule
Recall@$k$ & Share of the gold EntitySets that appear in the top $k$ results. \\
MRR & Mean of $1/\text{rank}$ of the first correct result. \\
nDCG@10 & Ranking quality of the top 10, giving more weight to higher ranks. \\
Completion rate & Share of agent runs that finish the task with valid calls. \\
\bottomrule
\end{tabular}
\end{table}

\vspace{1mm}\noindent \textbf{Enterprise systems and SAP.}
Companies run their core business tasks, such as purchasing, sales, and warehousing, in enterprise resource planning (ERP) software. SAP S/4HANA is one such ERP system. It gives other programs access to its data through web services, also called APIs. SAP publishes the descriptions of these services in a public catalogue, the SAP Business Accelerator Hub. We use only these public descriptions, so no running SAP system is needed.

\vspace{1mm}\noindent \textbf{OData services.}
The Open Data Protocol (OData)\footnote{OASIS OData Version 4.0, \url{https://docs.oasis-open.org/odata/odata/v4.0/odata-v4.0-part1-protocol.html}} is a standard for web services that read and write data over HTTP. Each OData service publishes a machine-readable description of itself, called its \texttt{\$metadata} document. An \emph{EntitySet} is a named collection of business records, similar to a database table, for example \texttt{PurchaseOrder}. Its \emph{EntityType} defines the structure of a record, and each field of a record is a \emph{Property}, for example \texttt{Supplier}. A \emph{NavigationProperty} links one EntitySet to another and states the cardinality: one purchase order has many purchase-order items (1:N). Some properties accept only codes from a fixed list, which SAP calls \emph{value help}. For example, the purchase-order type must be \texttt{NB} (standard order), \texttt{UB} (stock transport order), or \texttt{FO} (framework order). A client reads records with a GET call. It creates a record with a POST call, whose body (the \emph{payload}) holds the property values. The \texttt{\$metadata} document says what data exists, but not in which order a business uses it.

\vspace{1mm}\noindent \textbf{BPMN processes.}
Business Process Model and Notation (BPMN 2.0)\footnote{OMG Business Process Model and Notation 2.0, \url{https://www.omg.org/spec/BPMN/2.0/}} is a standard for describing business processes. A BPMN model is a flow chart of \emph{activities} (tasks) connected by \emph{sequence flows}. It is stored as XML, so programs can read it. A purchasing process, for example, may run \emph{create purchase order}, then \emph{post goods receipt}, then \emph{enter supplier invoice}. A BPMN model says in which order tasks happen, but not which data each task uses. Our method links the two standards: each activity is connected to the EntitySets it reads or writes.

\vspace{1mm}\noindent \textbf{Property graphs.}
A property graph is a kind of knowledge graph. It stores data as nodes and directed edges, and both can carry a type label and key-value attributes. Cypher~\cite{Francis:2018:SIGMOD:cypher} is a query language for such graphs. A Cypher query can follow a path of bounded length, for example ``from this activity, follow up to three \emph{next} edges''. Such bounded queries have predictable cost. We use FalkorDB\footnote{\url{https://docs.falkordb.com}}, an open-source graph database that runs Cypher.

\vspace{1mm}\noindent \textbf{Dense retrieval.}
An embedding model turns a text into a vector, so that texts with similar meaning get nearby vectors. Dense retrieval embeds the query and returns the $k$ stored items with the closest vectors. We use the FAISS library~\cite{douze2024faiss} for this search. Dense retrieval works well when the query describes the target. It has no way to find a target that the query does not describe, such as the next step of a process.

\vspace{1mm}\noindent \textbf{LLM agents.}
An LLM agent is a large language model that runs in a loop. It reasons about the next step, calls a tool, reads the result, and repeats until the task is done. This pattern is called ReAct~\cite{yao:2023:ICLR:react}. In this setting, \emph{hallucination} means that the agent invents an EntitySet, a key, or a code that does not exist in the service description, for example the code \texttt{NORMAL} instead of \texttt{NB}. We call any such call an \emph{invalid call}.

\begin{figure*}[t]
    \centering
    \includegraphics[width=\linewidth, trim={.0cm 0.0cm .0cm 0.0cm}, clip]{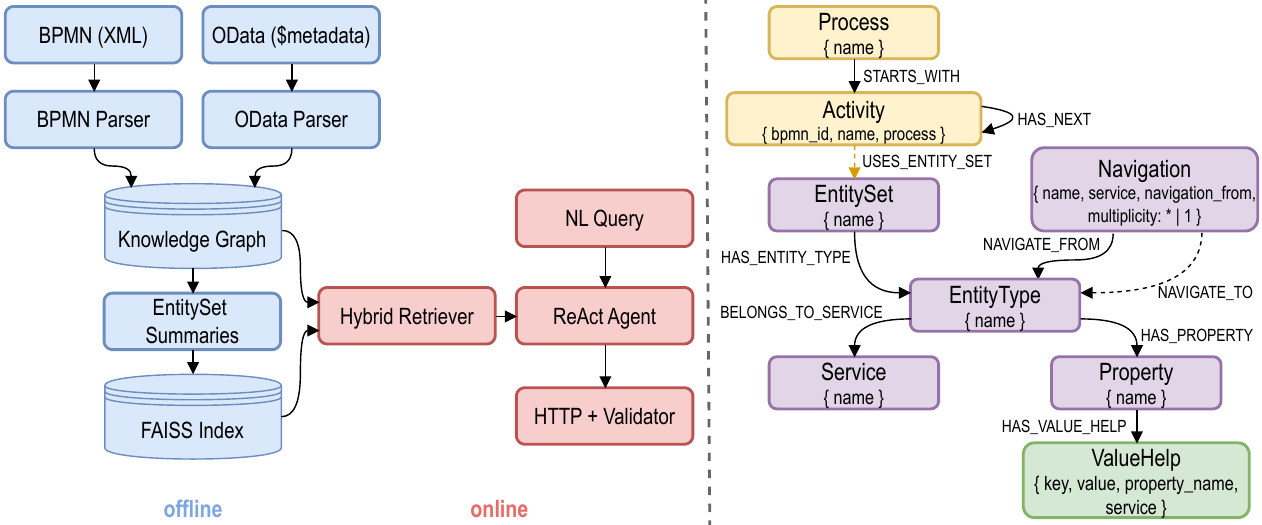}
    \vspace{-5mm}
    \caption{(Left) End-to-end architecture: offline ingestion and indexing feed an online hybrid
    retriever and an execution-capable ReAct agent.
    (Right) Property-graph schema: OData structural nodes (purple); BPMN process nodes (orange);
    \texttt{ValueHelp} enum node (green).}
    \label{fig:arch}
\end{figure*}

\section{Related Work}
\label{sec:relwork}

\subsection{Retrieval-augmented service discovery}

Pesl et al.~\cite{Pesl:IEEE-TSC:2026} compare chunking strategies for OpenAPI specifications, including ChunkRAG~\cite{singh2025:chunkrag}. They also introduce SOCBench-D, a benchmark of 550 queries across 11 GICS sectors. Their best method, endpoint splitting with summaries, is our chunk-RAG baseline. They leave knowledge graphs to future work, and their Discovery Agent does not execute calls. Earlier work in the same line, such as CRAFT~\cite{pesl2025craft} and RestGPT~\cite{song2023restgpt}, combines retrieval with single-call execution. Their position papers~\cite{pesl2023uncovering,pesl2024compositio} frame LLM-based service composition as a wider research agenda. We extend this line to multi-step, payload-aware composition with explicit process context.

Expanding a knowledge graph around dense retrieval seeds is also common in KG-augmented RAG, for example in GraphRAG~\cite{edge2024graphrag}. Our claim is narrower. In our experiments, generic graph structure over service metadata does \emph{not} improve retrieval. The gain comes from typed process-sequence edges. Graphs built only from extracted entities do not have such edges.

\subsection{Classical composition and knowledge graphs for services}
\label{sec:relwork-classical}

In service computing~\cite{Khan:2022:TSC,Khan:2025:SCC}, automated service composition is decades older than LLMs. Semantic web service frameworks such as OWL-S~\cite{martin2007owls}, WSMO~\cite{roman2005wsmo}, and SAWSDL~\cite{kopecky2007sawsdl} describe service capabilities in machine-readable form. They need a lot of manual modeling, and this cost is one reason given~\cite{Pesl:IEEE-TSC:2026} for preferring chunk-RAG over knowledge graphs. Other work treats composition as AI planning. HTN planners such as SHOP2~\cite{sirin2004htn} turn goals into executable service chains; see~\cite{rao2004survey} for a survey. QoS-aware methods~\cite{zeng2004qos} optimize non-functional constraints. Recent non-LLM methods learn service representations directly, for example with feature mining and contrastive learning~\cite{xu2025restful}. Our process edges look like the workflow skeletons that planners use. The difference is that we derive them from BPMN reference processes and give them to an LLM as retrieval context, not to a symbolic planner.

Our approach reduces this manual work but does not remove it. OWL-S and WSMO need annotations for every service in a catalogue. We load OData \texttt{\$metadata} automatically and write by hand only the BPMN layer and a small activity-to-EntitySet mapping (Sec.~\ref{sec:discussion-mapping}).

\subsection{LLM tool use and agent frameworks}

ReAct~\cite{yao:2023:ICLR:react} introduced the pattern of interleaving reasoning and acting. Several works connect LLMs to large API collections. Toolformer~\cite{schick2023toolformer} learns tool calls through self-supervision. Gorilla~\cite{patil2024gorilla} retrieves and invokes APIs from large documentation sets. ToolLLM~\cite{qin2024toolllm}, AnyTool~\cite{du2024anytool}, and API-Bank~\cite{li2023apibank} evaluate tool calling over broad, loosely typed REST APIs, usually from free-form documentation and for single calls or short chains. None of them model the structural rules, such as fixed value lists or navigation cardinality, that decide whether a payload is valid. In our system, every retrieved endpoint comes with these rules, and the mock execution layer enforces them.

\subsection{Business-process and API integration}
Graph transformers improve web API recommendation~\cite{zheng2024motif}, and LLM-based representations help recommend services with little usage history~\cite{rong2024llmcoldstart}. LLMs also generate TOSCA service compositions from natural-language requirements~\cite{meflah2025tosca} and drive business-process discovery~\cite{Lindenberg2026:ICSOC}. Goretti et al.~\cite{Valerio2026:ICSOC} enforce usage policies during process discovery with a Trusted Execution Environment. We focus on accurate service matching and on the schema and process context an agent needs before it can execute a call.

Work that links business-process models to executable services has mostly focused on modeling. Examples include deriving orchestrations from BPMN~\cite{ouyang2009bpmn} and industrial tools such as Camunda Connectors\footnote{\url{https://camunda.com/platform/modeler/connectors/}} and SAP Signavio reference processes\footnote{\url{https://www.signavio.com}}. This link has rarely been used as the basis for retrieval and execution in LLM-based discovery. Work on generating BPMN from text~\cite{nivon2024bpmngen} may one day automate the process layer that we currently write by hand. In our system, the activity-to-endpoint mapping is a first-class graph edge that the retriever uses directly. To our knowledge, no earlier work has evaluated process-chain expansion in retrieval-augmented service discovery. Our benchmark provides a repeatable setting for doing so.

\section{Approach}
\label{sec:approach}

\begin{figure*}[t]
\centering
  \begin{lstlisting}[style=code, basicstyle=\ttfamily\scriptsize]
  THINK  user wants PO + goods receipt; need entity sets
  ACT    RETRIEVE("create purchase order then goods receipt")
  OBS    specs: PurchaseOrder { required: Supplier, PurchaseOrderType;
          PurchaseOrderType value-help: [ "NB", "UB", "FO" ];
          nav: PurchaseOrderItem (1:N) }, MaterialDocumentHeader { ... }
        process: Create purchase order -> Post goods receipt
  THINK  PurchaseOrderType must come from the value-help list -> "NB"
  ACT    POST PurchaseOrder { Supplier:"4711", PurchaseOrderType:"NB",
          PurchaseOrderItem:[ { Material:"M-001", Quantity:5 } ] }
  OBS    201 -> { PurchasingDocument: "4500001234" }
  THINK  next activity in chain is goods receipt
  ACT    POST MaterialDocumentHeader
          { ReferenceDocument:"4500001234", DocumentType:"WE" }  -> 201
\end{lstlisting}

\caption{Agent trace for a query that creates a purchase order and then posts the goods receipt. THINK is the agent's reasoning, ACT a tool call, and OBS the tool result. The retrieved value-help list stops the agent from inventing a \texttt{PurchaseOrderType} code. Process expansion shows that \texttt{MaterialDocumentHeader} comes next.}
\label{fig:agent-trace}
\end{figure*}

\subsection{System Overview}

Fig.~\ref{fig:arch}(left) shows the two stages of our system. In the offline stage, the ingestion pipeline reads OData \texttt{\$metadata} and BPMN 2.0 files and writes the nodes and edges to the graph database. It also writes a short text summary for each EntitySet and indexes the summaries with FAISS~\cite{douze2024faiss}.

In the online stage, the hybrid retriever takes the top-$k$ dense matches as seeds and expands them along the process chain with a bounded Cypher~\cite{Francis:2018:SIGMOD:cypher} query. A ReAct~\cite{yao:2023:ICLR:react} agent then issues GET and POST calls, which a mock HTTP layer validates.

Fig.~\ref{fig:arch}(right) shows the graph schema. Eight node labels represent OData and BPMN structure, and nine edge types connect them. Two design choices matter for execution. First, we store cardinality on \texttt{Navigation} nodes. This keeps graph queries simple and shows the agent which relations are 1:N and which are N:1. Second, \texttt{ValueHelp} is its own node and holds the list of valid codes. The agent can therefore get the valid codes with one bounded graph lookup, instead of relying on a loose embedding match.

\subsection{Offline: Ingestion Pipeline}

\vspace{1mm}\noindent \textbf{OData parser.}
The OData parser reads the EntityTypes, EntitySets, and SAP annotations of each \texttt{\$metadata} document. Each EntityType becomes a node, and its Properties become child nodes linked by \texttt{HAS\_PROPERTY}. Each NavigationProperty becomes a \texttt{Navigation} node. It links to its source and target EntityTypes through \texttt{NAVIGATE\_FROM} and \texttt{NAVIGATE\_TO} and stores its cardinality (one or many). ValueHelp annotations become nodes, linked by \texttt{HAS\_VALUE\_HELP}, that hold the explicit code lists. Table~\ref{tab:dataset} gives the counts.

\vspace{1mm}\noindent \textbf{BPMN parser.}
The BPMN parser extracts the activities of each process and builds \texttt{HAS\_NEXT} edges from the sequence flow. We ignore gateways, events, and pools. For our purpose, the value of the BPMN layer is the order of activities, not the full execution semantics.

\vspace{1mm}\noindent \textbf{Activity-to-EntitySet mapping.}
A short, hand-written mapping links each activity to the EntitySets it reads or writes (e.g., \emph{create purchase order} $\to$ \texttt{PurchaseOrder}, \texttt{PurchaseOrderItem}). It covers 14 processes, 51 task activities, and 45 \texttt{\seqsplit{USES\_ENTITY\_SET}} edges. The mapping is small enough to review in full, yet it drives the retrieval gains reported in Sec.~\ref{sec:eval}. We leave ten activities unmapped on purpose, because no service in the catalogue covers their systems (e.g., recruiting, payroll, outbound delivery).

\vspace{1mm}\noindent \textbf{EntitySet summaries.}
For the embeddings, we build a short text summary of each EntitySet. It joins the EntitySet label, the EntityType description, and the Property names. This follows the ``summary'' chunking approach of~\cite{Pesl:IEEE-TSC:2026}, but at the EntitySet level instead of the endpoint level.

\subsection{Online: Retrieval and Execution}
\label{sec:approach-retrieval}

\vspace{1mm}\noindent \textbf{Hybrid retrieval.}
The retriever has two branches. The dense branch runs a top-$k$ search over the EntitySet summaries and returns a set $S$ of likely matches. For each $e \in S$, the process branch runs a bounded graph query that finds downstream EntitySets: it goes from $e$ to the activities that use it, follows \texttt{HAS\_NEXT} to later activities, and collects the EntitySets those activities use. The retriever keeps the dense order, places each new entity after its seed, and removes duplicates. It also returns the cardinality of each navigation, so the agent knows how to shape the payload.

\vspace{1mm}\noindent \textbf{Execution agent.}
A ReAct agent, built with LangGraph\footnote{\url{https://langchain-ai.github.io/langgraph/}}, has three tools: (a) hybrid retrieval, which returns the full typed specification (required keys, navigation cardinalities, value-help lists, and process-chain information); (b) GET; and (c) POST. A mock HTTP layer validates every call. All grounding context arrives in one retrieval call, which keeps tool round-trips low for local 30B models. Two mechanisms reduce hallucination. First, POST rejects payloads with missing keys or invalid codes, so the agent can retry. Second, navigation cardinalities tell the agent to send 1:N relations as nested arrays. Fig.~\ref{fig:agent-trace} shows both mechanisms.

\vspace{1mm}\noindent \textbf{Worked example.}
\label{sec:approach-example}
Consider the query: \emph{``Create a purchase order with a valid type code for supplier 4711, then post the goods receipt.''} A semantic retriever over endpoint summaries finds \texttt{PurchaseOrder}, but it may miss \texttt{PurchaseOrderItem} because the query never says ``item''. The valid type codes exist only in ValueHelp, so an LLM that sees only summaries may output a plausible but invalid code such as \texttt{NORMAL}. It also does not know that \texttt{\seqsplit{MaterialDocumentHeader}} must follow \texttt{PurchaseOrder}. Fig.~\ref{fig:agent-trace} shows how our system closes each gap: the dense seed finds \texttt{PurchaseOrder}, process expansion adds \texttt{\seqsplit{MaterialDocumentHeader}}, and the retrieved specification gives the 1:N item navigation and the value-help list.

\section{Experimental Results}
\label{sec:eval}

We evaluate our approach with a construction-fidelity audit and five research questions (RQs). Each one targets a contribution from Sec.~\ref{sec:introduction}.

\vspace{1mm}\noindent \textbf{Fidelity audit (contribution~(a)):} Does the graph match the source OData \texttt{\$metadata} and BPMN files (Sec.~\ref{sec:eval-fidelity})?

\vspace{1mm}\noindent \textbf{RQ1 (retrieval quality; contribution~(b)):} Does process-aware hybrid retrieval find more correct endpoints than dense-only and graph-only retrieval? RQ1 uses each benchmark query word for word, so it measures retrieval without any agent behaviour.

\vspace{1mm}\noindent \textbf{RQ2 (ablation; contributions~(a),(b)):} Which graph component causes the gain?

\vspace{1mm}\noindent \textbf{RQ3 (end-to-end agent; contribution~(c)):} Can the agent complete multi-step tasks without invalid calls? RQ3 measures the full loop, so retrieval errors and execution errors are mixed. An oracle-retrieval test separates them.

\vspace{1mm}\noindent \textbf{RQ4 (hallucination elicitation; contribution~(c)):} Does value-help grounding stop invented codes and keys when a query needs a fixed-value code?

\vspace{1mm}\noindent \textbf{RQ5 (external generalization; limits of contribution~(b)):} Does the gain carry over to OpenAPI services that have no business-process layer? This is a falsification test: if our explanation is right, there should be \emph{no} gain.

\subsection{Datasets}
\label{sec:eval-data}

\begin{table}[!htb]
\centering
\caption{OData-Proc dataset statistics. Structural counts are distinct elements after removing $29$ duplicate source rows (Sec.~\ref{sec:eval-fidelity}). Most EntityTypes are complex, nested, or parameter types that are never exposed as a top-level collection. The RQ1 process-chain probe ($n=74$) is exactly the process-chain goal type.}
\label{tab:dataset}
\footnotesize
\setlength{\tabcolsep}{2pt}
\begin{tabular}{@{}lr@{\hspace{6pt}}lr@{}}
    \toprule
    \textbf{Element} & \textbf{Count} & \textbf{Queries} & \textbf{Count} \\
    \midrule
    Source services          &     42 & Total                 & 288 \\
    EntitySets               &    132 & \quad Single entity        & 130 \\
    EntityTypes              &  1,270 & \quad Process-chain goal   &  74 \\
    Properties               &  8,481 & \quad Navigation chain     &  67 \\
    Navigation edges         &  1,499 & \quad Multi-step workflow  &  15 \\
    BPMN processes           &     14 & \quad Value-help dep.      &   2 \\
    BPMN task activities     &     51 & Train / test          & 150/138 \\
    Mapping edges            &     45 & RQ3 queries &  81 \\
    \bottomrule
\end{tabular}
\end{table}

Our main benchmark, \emph{OData-Proc}, combines four public OData corpora (SAP Business Accelerator Hub, TripPin, Northwind, and Microsoft Graph) with 14 BPMN reference processes (Table~\ref{tab:dataset}). It uses only openly published metadata, so no access to a proprietary system is needed. Eight of the process files cover SAP reference scenarios such as Warehouse Inbound Processing and Sell from Stock. We converted these files to BPMN~2.0 from an earlier RDF format and kept the activity names, order, and entity-set mappings.

There are five query types. A \emph{single-entity} query asks for one EntitySet. A \emph{navigation-chain} query needs EntitySets linked by navigation. A \emph{process-chain goal} states a business goal whose downstream EntitySets share few words with the query. A \emph{multi-step workflow} asks for several explicit steps, and a \emph{value-help dependent} query needs a fixed code. The benchmark has $288$ queries, split deterministically into $150$ training and $138$ test queries. We used the training set only to tune prompt wording and the retriever settings, and did not look at the test set until the final evaluation. All $74$ process-chain probes are in the test split. The value-help category has only $2$ queries on purpose: we test code grounding on a separate $56$-query elicitation probe with adversarial wording, so that it does not inflate the main retrieval scores. The RQ3 execution experiments use the $81$ test queries of the four multi-step types that target the SAP corpus. A mock HTTP layer validates payloads against the OData metadata, so the experiments need no live backend. For external generalization, we sample 110 SOCBench-D queries, 10 from each GICS sector~\cite{Pesl:IEEE-TSC:2026}.

The first version of the benchmark had $80$ queries labelled by one annotator. We grew it by generating $251$ candidate queries from the catalogue and the process graph, each with machine-suggested labels, and having two annotators review every candidate independently: a human author and a large language model (Claude, Anthropic), whose per-candidate decisions we release. They agreed on whether to keep $250$ of the $251$ candidates (Cohen's $\kappa = 0.665$) and on all $415$ (query, EntitySet) labels of the candidates both kept ($\kappa = 1.000$). Both saw the suggested labels, so this measures agreement on corrections, not independent labelling. The merge added $208$ queries to the main benchmark and $41$ to the elicitation probe (full protocol in Supplementary Sec.~S2).

\subsection{Experimental Setup}
\label{sec:eval-setup}

We ran all experiments on an A100 GPU (40\,GB VRAM). They can also be reproduced on a workstation with 32\,GB RAM and no GPU. FalkorDB stores the graph, and Ollama\footnote{\url{https://ollama.com}} hosts the LLMs and embedding models locally. The primary LLM is \texttt{\seqsplit{qwen3-coder:30b}} (30B, tool-tuned). We use \texttt{\seqsplit{qwen3.6}} (36B, general) and \texttt{\seqsplit{gemma4:31b}} for sensitivity analysis. The main embedding model is \texttt{\seqsplit{qwen3-embedding:8b}}. As a control, we also use the smaller \texttt{\seqsplit{sentence-transformers/all-mpnet-base-v2}}. Table~\ref{tab:glossary} defines the metrics.

Unless stated otherwise, every $p$ value in this paper comes from a paired two-sided Wilcoxon signed-rank test over per-query scores; where we compare all three retrievers at once, we use the Friedman test and call it the overall test. Both are standard tests for paired results that do not assume a normal distribution.

\subsection{Graph-Construction Fidelity}
\label{sec:eval-fidelity}

All later results depend on the graph, so we first check that it matches its sources. An independent checker, written separately from the ingestion pipeline, reads the source OData and BPMN files again and compares them with the graph. For each structural class we measure \emph{coverage} (the share of source elements found in the graph) and \emph{exactness} (the share of graph elements whose attributes match the source). For EntitySets, EntityTypes, Properties, Navigation edges with cardinality, ValueHelp codes, BPMN activities, and activity order, both coverage and exactness are $1.000$. The few differences come from the sources (number formats and duplicate rows), not from ingestion. The audit does not cover the hand-written mapping edges, because no independent source exists for them; we assess them with the provenance split in Sec.~\ref{sec:eval-rq1}. Supplementary Sec.~S3 gives the full audit.

\subsection{RQ1: Retrieval quality}
\label{sec:eval-rq1}

\begin{table*}[t]
\centering
\caption{
RQ1 retrieval quality on the OData-Proc test split ($n=138$). Hybrid is our process-aware method; Dense is FAISS-only retrieval; Graph is graph-only retrieval. The lower block shows the process-chain probe subset ($n=74$); there, Hybrid beats Dense at Recall@3 with $p=0.037$ (paired test).
}
\label{tab:rq1}

\footnotesize
\setlength{\tabcolsep}{3pt}
\renewcommand{\arraystretch}{1.05}
\begin{tabular}{lcccccccr}
    \toprule
    \textbf{Cond.} & \textbf{R@1} & \textbf{R@3} & \textbf{R@5} &
    \textbf{R@10} & \textbf{MRR} & \textbf{nDCG@10} &
    \textbf{lat.\,(ms)} \\
    \midrule
    \multicolumn{8}{l}{\emph{Full set ($n=138$)}} \\
    Dense              & 0.472 & 0.787 & 0.870 & 0.928 & \textbf{0.902} & 0.868 & 202.9 \\
    Graph              & 0.132 & 0.292 & 0.394 & 0.617 & 0.352 & 0.397 &  15.6 \\
    \textbf{Hybrid} & 0.472 & \textbf{0.810} & \textbf{0.885} & \textbf{0.945} & 0.893 & \textbf{0.875} & 208.4 \\
    \midrule
    \multicolumn{8}{l}{\emph{Process-chain probe ($n=74$)}} \\
    Dense              & 0.318 & 0.675 & 0.788 & 0.869 & \textbf{0.889} & 0.809 & 203.9 \\
    Graph              & 0.012 & 0.068 & 0.142 & 0.419 & 0.181 & 0.222 &  15.4 \\
    \textbf{Hybrid} & 0.318 & \textbf{0.732} & \textbf{0.822} & \textbf{0.901} & 0.877 & \textbf{0.837} & 199.9 \\
    \bottomrule
    \end{tabular}
\end{table*}

We compare Dense (FAISS only), Graph (graph only), and Hybrid (our method) on the 138 test queries of OData-Proc. Table~\ref{tab:rq1} reports Recall@$\{1,3,5,10\}$, MRR, nDCG@10, and latency against the labelled gold EntitySets.

As expected, the gain appears only where there is a process signal. On the process-chain probe ($n=74$), the downstream entities share few words with the query. Here Hybrid improves Recall@3 by $+5.7$ points over Dense ($0.732$ vs.\ $0.675$; $p=0.037$). The gap narrows to $+3.4$ points at $k=5$ ($p=0.13$) as dense retrieval catches up, and widens again at $k=10$ ($+3.2$ points, $p=0.018$). On the $64$ non-probe queries, Hybrid stays close to Dense (Recall@3 $0.901$ vs.\ $0.917$). The small loss comes from expansion sometimes pushing out a correct dense result. Over the full test split, Hybrid gains $+2.3$ points at $k=3$ ($0.810$ vs.\ $0.787$; overall test $p = 1.7\times10^{-31}$, $n=138$). This gain comes entirely from the probe, which makes up more than half of the test split. The effect is smaller than in the earlier 80-query version of the benchmark ($+15.0$ probe Recall@3 points on $n=20$). Many of the new probe queries were reworded to name their business process explicitly, which gives dense retrieval more words to match.

MRR barely changes ($0.893$ for Hybrid vs.\ $0.902$ for Dense). Expansion places new entities after the dense seed, so its benefit is to fill ranks 2 and 3 with downstream entities that dense retrieval cannot reach. At $k=3$, expansion adds $32$ of the $304$ gold entities ($28$ of them in the probe) and pushes $16$ dense golds out of the top three. Graph alone performs poorly: without embedding seeds, a graph-only search cannot tell candidates apart. This supports the hybrid design. Supplementary Sec.~S4 gives the full per-type breakdown and shows that, whenever both entities of a required pair are retrieved, the context passed to the agent is complete.

\vspace{1mm}\noindent \textbf{Ground-truth provenance split.}
The same author wrote the activity-to-EntitySet mapping and the query labels. To control for this circularity, we split the test queries by where their labels come from. Labels for 59 queries follow from the OData metadata alone (no process chain needed). The other 79 queries need the mapping. On the metadata-only queries, Hybrid and Dense tie at Recall@1 ($0.675$). Hybrid is slightly lower at Recall@3 and Recall@5 ($0.915$ vs.\ $0.932$; $0.958$ vs.\ $0.975$), because expansion sometimes pushes out a late gold. All of the gain is in the mapping-dependent subset (Recall@3 $0.678 \to 0.732$, nDCG@10 $0.815 \to 0.844$). Performance therefore depends on a correct mapping.

\vspace{1mm}\noindent \textbf{Validity and scope.}
\emph{(i)~Construction.} The ground-truth chains of the probe come from the hand-written mapping, and the queries are written to avoid word overlap. This favours Hybrid by design. The probe therefore tests whether curated process edges make otherwise unreachable entities retrievable, not whether such edges exist by default.
\emph{(ii)~Multiple comparisons.} The overall test is clear ($p=1.7\times10^{-31}$). We ran two paired tests on the probe, so a stricter threshold of $0.025$ applies to each (Holm correction); the main probe result ($p=0.037$) does not meet it. With the smaller mpnet embedding, Hybrid still beats Dense on the probe at every $k$, but none of these differences is significant (all $p \geq 0.19$). We therefore read the probe-level gain as consistent in direction but not confirmed at its current effect size.
\emph{(iii)~Baseline choice.} Dense already implements Pesl's ``summary'' chunking at the EntitySet level, so it stands in for chunk-RAG on OData. Porting their token-overlap retriever to OData would risk measuring the port, so we evaluate it on its native benchmark, SOCBench-D (RQ5).

\subsection{RQ2: Ablation}
\label{sec:eval-rq2}

We remove Hybrid components one at a time. Removing \texttt{\seqsplit{USES\_ENTITY\_SET}} and \texttt{HAS\_NEXT}, or all graph branches, turns Hybrid into Dense: the metrics match the Dense rows of Table~\ref{tab:rq1} at every $k$. The Recall@3 gain therefore comes from the process edges, not from the other graph elements. We do not ablate ValueHelp here, because it is used to build payloads, not to rank results; RQ3 and RQ4 test it.

\vspace{1mm}\noindent \textbf{Separating the two process-edge types.}
Are the $45$ hand-written mapping edges enough on their own, or does BPMN sequence order also help? \emph{Mapping-only} keeps \texttt{\seqsplit{USES\_ENTITY\_SET}} but removes \texttt{HAS\_NEXT}, so expansion finds the entities used by the seed activity but cannot follow the chain. \emph{Sequence-only} keeps \texttt{HAS\_NEXT} but removes the mapping, which cuts the path from entities to activities. Both stay at the Dense level (probe Recall@3 $0.673$ and $0.675$; full split $0.786$ and $0.787$). Full Hybrid beats mapping-only on the probe by $+5.9$ Recall@3 points ($p=0.028$; $+2.6$ at $k=5$, $p=0.33$). Both edge types are therefore needed: the mapping ties entities to activities, and \texttt{HAS\_NEXT} carries the expansion to later activities.

\subsection{RQ3: End-to-end agent}
\label{sec:eval-rq3}

\begin{table}[!htb]
\centering
\caption{
RQ3: mock-execution agent on the $81$ multi-step test queries ($3$ runs per query, primary LLM). ``Calls/run'' is the mean number of tool calls per run, and ``Invalid/run'' the mean number of invalid calls per run (\texttt{GET \$metadata} requests are not counted). No run timed out. Completion difference $+0.284$, $p<0.001$ (paired test, $n=81$). Results on an earlier 26-query subset and with other LLMs are in Supplementary Sec.~S5.
}
\label{tab:rq3}

\footnotesize
\setlength{\tabcolsep}{3pt}
\begin{tabular}{lccccc}
    \toprule
    \textbf{Cond.} & \textbf{Runs} & \textbf{Compl.} & \textbf{Calls/run} &
    \textbf{Invalid/run} & \textbf{Lat.\,(s)} \\
    \midrule
    ChunkRAG       & 243 & 0.103 & 0.89 & 0.107 & 5.3 \\
    \textbf{Ours}  & 243 & \textbf{0.387} & 2.45 & 0.103 & 11.5 \\
    \bottomrule
\end{tabular}
\end{table}

Table~\ref{tab:rq3} shows agent results on the $81$ multi-step test queries ($20$ navigation chains, $58$ process-chain goals, $2$ multi-step workflows, and $1$ value-help query). Both conditions share the same mock GET/POST layer and validator. ChunkRAG~\cite{singh2025:chunkrag} combines the chunk-RAG retriever of~\cite{Pesl:IEEE-TSC:2026} with our execution layer and returns entity set, service, and property names. Our agent's retrieval returns the full typed specification: required keys, navigation cardinality, value-help code lists, and process-chain details. ChunkRAG therefore cannot see value-help lists, by design (Supplementary Sec.~S1 compares the tools). We run each query three times ($243$ runs per condition).

\vspace{1mm}\noindent \textbf{Scope and circularity.}
These experiments use no live backend. The mock validator checks payloads against the same metadata and ValueHelp nodes that we used to build the graph, so our agent makes an invalid call only when it ignores or misuses the retrieved context. The invalid-call metric therefore tests whether grounding closes this failure mode by construction, not how the agent performs against a live service (Sec.~\ref{sec:discussion-threats}).

\vspace{1mm}\noindent \textbf{Completion rate.}
Our agent completes far more tasks than ChunkRAG ($0.387$ vs.\ $0.103$, $p<0.001$). The gain holds in every type with more than two queries: navigation chains ($0.517$ vs.\ $0.100$) and process-chain goals ($0.316$ vs.\ $0.103$). Two factors explain it. First, ChunkRAG often stops before acting: it makes no GET or POST call in $173$ of $243$ runs (ours: $78$). Second, when in-loop discovery returns all gold entities, our agent completes $0.489$ of runs, compared with $0.175$ for ChunkRAG. Retrieval misses still cause many failures for both methods: $100$ of $218$ ChunkRAG failures and $57$ of $149$ of ours. Our agent is slower ($11.5$ vs.\ $5.3$ seconds per run), mainly because it makes more tool calls ($2.45$ vs.\ $0.89$).

An earlier 26-query subset, written before the benchmark grew, pointed the other way: ChunkRAG completed more tasks ($0.295$ vs.\ $0.192$), but the difference was not significant ($p=0.46$). On that subset, a wrong dense seed was the main risk, because process expansion then adds the wrong chain. We base our conclusions on the larger test set and report the subset in Supplementary Sec.~S5.

\vspace{1mm}\noindent \textbf{Invalid-call rate.}
Both agents make about the same number of invalid calls per run ($0.103$ for ours vs.\ $0.107$ for ChunkRAG; $25$ vs.\ $26$ in total). Neither invents a code. Our agent writes much more often ($55$ vs.\ $22$ runs with a POST), so the per-run rates hide this difference. Most of our invalid calls read or write an entity set that the service does not have ($16$ of $25$). Most of ChunkRAG's invalid calls use a key that does not exist ($21$ of $26$). These queries rarely need fixed-value codes, so the elicitation probe in Sec.~\ref{sec:eval-rq3b} tests this case directly.

\vspace{1mm}\noindent \textbf{Separating execution from retrieval.}
To separate the two failure sources, we ran an oracle condition on the 26-query subset: each agent receives the gold specification for each query (perfect retrieval). Completion stays low for both agents (ours $0.256$, ChunkRAG $0.359$), so execution by the 30B LLM, not only retrieval, limits completion. Payload construction still separates the two conditions. With the same gold context, ChunkRAG issues $15$ schema-invalid writes, including $6$ invented codes, while ours issues $3$ and invents none. Supplementary Sec.~S5 gives the full oracle results, a seed-verification variant, and the sensitivity runs with \texttt{gemma4:31b} and \texttt{qwen3.6}.

\subsection{RQ4: Hallucination elicitation}
\label{sec:eval-rq3b}

\begin{table}[!htb]
\centering
\caption{
RQ4: code elicitation on the $56$-query probe ($3$ runs per query; $168$ per condition; primary LLM). ``POST runs'': runs with at least one POST. ``Inv.~runs'': POST runs with an invalid call. ``Invalid'': all invalid calls (entity set, key, or code). ``Codes'': invented codes. ChunkRAG+VH adds the value-help code lists as plain text to the retrieved chunks. No run timed out.
}
\label{tab:rq3b}

\footnotesize
\setlength{\tabcolsep}{2.5pt}
\renewcommand{\arraystretch}{1.05}
\begin{tabular}{lcccc}
    \toprule
    \textbf{Cond.} & \textbf{POST runs} &
    \textbf{Inv.\ runs} & \textbf{Invalid} & \textbf{Codes} \\
    \midrule
    ChunkRAG          & 127 & 100/127 & 218 & 60 \\
    ChunkRAG+VH & 128 & 59/128 & 120 & \textbf{0} \\
    \textbf{Ours} & 101 & \textbf{3/101} & \textbf{3} & \textbf{0} \\
    \bottomrule
\end{tabular}
\end{table}

We build a 56-query elicitation probe, separate from the main benchmark. It started with $15$ hand-written queries; the two-annotator protocol (Sec.~\ref{sec:eval-data}) added $41$ more. Each query implies a specific SAP fixed-value code without naming it. For example, ``a \emph{standard} purchase order'' implies \texttt{PurchaseOrderType=NB}. The codes span three EntitySets that accept POST and use real SAP value-help domains. The ingested \texttt{\$metadata} has no machine-readable value lists, so we enter the relevant domains by hand. The graph and the mock validator use the same values, so the circularity note in Sec.~\ref{sec:eval-rq3} applies.

Table~\ref{tab:rq3b} shows the result. Because the conditions issue POSTs at different rates, we report invalid-call rates over runs that issue a POST. ChunkRAG makes an invalid call in $100$ of $127$ POST runs ($0.79$, 95\% CI $[0.71, 0.85]$), with $218$ invalid calls in total. Sixty of these are invented codes, such as \texttt{DA} or \texttt{EX WORKS} instead of the Incoterms codes \texttt{DAP} and \texttt{EXW}, \texttt{RET} instead of the returns code \texttt{RE}, and \texttt{IV} instead of the invoice code \texttt{F2}. The most common invented code is the purchasing code \texttt{NB} reused as a sales-order type ($19$ times), so pre-training is not reliable even for well-known codes. Our method makes an invalid call in $3$ of its $101$ POST runs ($0.03$, 95\% CI $[0.01, 0.08]$) and invents no code. Two of its three invalid calls put a header key (\texttt{ShippingCondition}) on a sales-order item, and one reads an entity set from the wrong service. Per query, our invalid-run rate is lower than ChunkRAG's ($p<0.001$, $n=56$).

One cost remains. In $45$ of our $168$ runs, the agent makes no tool call and replies in text only (ChunkRAG: $19$). This is a failure to use the tools, not a timeout, and it is not hidden partial execution, because these runs did nothing.

\vspace{1mm}\noindent \textbf{Symmetric baseline: ChunkRAG with inlined value-help.}
Plain ChunkRAG cannot see value-help lists. To separate ``the value list was visible'' from ``a typed graph carried it'', we add a third condition that pastes the value-help code lists as plain text into ChunkRAG's retrieved chunks. Everything else stays the same. Pasting the lists closes the code class: ChunkRAG+VH invents no codes in its $128$ POST runs. Showing the value list is therefore enough to stop invented codes, whatever form it takes. It does not stop other invalid calls: $120$ remain in $59$ of the $128$ POST runs ($0.46$, 95\% CI $[0.38, 0.55]$), because flat text does not constrain the payload structure. With the full typed context, the rate falls to $0.03$ ($p<0.001$ per query against ChunkRAG+VH, $n=56$).

\subsection{RQ5: External generalization}
\label{sec:eval-rq4}

\begin{table}[!htb]
\centering
\caption{
RQ5: generalization on SOCBench-D (OpenAPI, no BPMN layer), pooled over 11 GICS sectors ($n=110$ queries). Hybrid vs.\ Tokens (paired tests): R@3 $+0.010$, $p=0.753$; R@5 $-0.021$, $p=0.437$; R@10 $-0.041$, $p=0.071$.
}
\label{tab:rq4}

\footnotesize
\setlength{\tabcolsep}{4pt}
\renewcommand{\arraystretch}{1.05}
\begin{tabular}{lccc}
    \toprule
    \textbf{Retriever} & \textbf{R@3} & \textbf{R@5} & \textbf{R@10} \\
    \midrule
    Tokens              & 0.297 & 0.413 & 0.614 \\
    Hybrid (no BPMN)    & 0.307 & 0.392 & 0.573 \\
    \bottomrule
\end{tabular}
\end{table}

Table~\ref{tab:rq4} shows the external test. We adapt the OData ingestion to OpenAPI 3.x and process the SOCBench-D specifications behind the 110 sampled queries (10 per GICS sector). SOCBench-D has no business-process layer, so the process branch of the Hybrid retriever is inactive, and only the weaker shared-parameter graph fallback applies. We compare against a token-overlap retriever (Tokens), which represents Pesl's chunk-RAG baseline.

Over all 110 queries, Hybrid and Tokens do not differ significantly at any $k$ (all $p\geq 0.07$). This is what we predicted: the gain depends on process-chain edges, not on graph structure in general. At $k=10$, Hybrid is slightly worse ($\Delta=-0.041$, $p=0.071$), because without process context, expansion adds some ranking noise. This small negative result further supports our explanation of the OData-Proc gain. Per-sector results (Supplementary Sec.~S7) are mixed and depend on how closely endpoint paths match the words in the queries.

\section{Discussion}
\label{sec:discussion}

\subsection{Mapping Effort and Scalability}
\label{sec:discussion-mapping}

The whole retrieval gain flows through the $45$ hand-written \texttt{\seqsplit{USES\_ENTITY\_SET}} edges, so their authoring cost is the main practical bottleneck. Three properties keep it low. First, the effort grows with the number of business processes, not with the size of the catalogue: the $42$ services, $132$ EntitySets, and $8{,}481$ properties were loaded with no manual work. Second, writing an edge only requires knowing which documents an activity reads or writes, which process owners already know; unlike OWL-S and WSMO annotations, it needs no ontology expertise. Third, the mapping is small enough to review in full.

We measured this cost directly. One author familiar with the catalogue re-wrote a random sample of ten mapped activities, covering $12$ of the $45$ edges, from scratch under a timer, following a fixed four-step procedure (Supplementary Sec.~S6). The sample took $205$ person-minutes (median $20.5$ minutes per activity, range $16$--$25$), and every re-written edge matched the released mapping. At the median rate, all $41$ mapped activities would take about $840$ person-minutes, or about two working days of focused effort.

For hundreds of processes, as in a full SAP Signavio library, manual authoring would still be costly. We therefore tested semi-automation, in which an LLM proposes edges and a human checks them. Given the catalogue and the ordered activities of each process, a local LLM (\texttt{qwen3-coder:30b}) proposes \texttt{\seqsplit{USES\_ENTITY\_SET}} edges. Against the $45$ hand-written edges, the proposals reach precision $0.79$ and recall $0.93$ (F1 $0.86$). Authoring the mapping thus becomes checking $53$ proposed edges instead of writing $45$ from scratch. The three missed edges show that domain review is still needed (Supplementary Sec.~S6).

\subsection{Threats to Validity}
\label{sec:discussion-threats}

\vspace{1mm}\noindent \textbf{Construct validity.}
The evaluation has a known circularity. The mock validator checks keys and codes against the same schema and ValueHelp nodes from which we build the graph. The low invalid-call results therefore show that grounding closes the failure mode \emph{by construction}; they do not measure performance against an independent target. In addition, the probe queries and the activity mapping have the same author. The provenance split, the fidelity audit, and the symmetric value-help baseline reduce these effects but do not remove them.

\vspace{1mm}\noindent \textbf{Internal validity.}
The benchmark has two-annotator ground truth over $288$ queries (Sec.~\ref{sec:eval-data}), but two concerns remain. First, the second annotator is a disclosed LLM that worked from the same machine-suggested labels as the human. The near-perfect label agreement therefore overstates how independent the labels are. Second, the probe-level retrieval effect is modest and does not survive Holm correction (Sec.~\ref{sec:eval-rq1}).

\vspace{1mm}\noindent \textbf{External validity.}
The gain depends entirely on process edges, so it only transfers to ecosystems that publish a process layer, or where one can be written. SOCBench-D confirms that there is no gain without one. SAP's reference processes are unusually well documented, and we have not yet shown a second domain with a native process layer. RQ5 thus defines the scope of the method: it targets structured enterprise ecosystems, and chunk-RAG remains a good fit for free-form APIs. Future work includes a second domain with a native process layer, support for GraphQL and gRPC, and automatic BPMN generation~\cite{nivon2024bpmngen} for domains without process models.

\section{Conclusion}
\label{sec:conclusion}

To call APIs reliably over several steps, an LLM agent must respect sequential and parametric constraints: process order, required keys, and valid code values. We meet this need by combining OData \texttt{\$metadata} and BPMN reference processes in one typed property graph. Hybrid retrieval combines dense embeddings with process-chain expansion, and a ReAct agent uses the resulting value-help and cardinality context to build payloads.
On process-chain queries, the method improves Recall@3 by $+5.7$ points ($p=0.037$, $n=74$), and it stays close to dense retrieval elsewhere. The ablation shows that the gain comes from process edges, not from generic graph structure. In the execution probe, typed context stops invented codes and almost all invalid keys: our method makes an invalid call in $3$ of its $101$ POST runs, while the baseline does so in $100$ of $127$, with $60$ invented codes. On multi-step test queries, our agent also completes more tasks than the baseline ($0.387$ vs.\ $0.103$), although completion remains low for both and the system is not yet ready for deployment. The evaluation is also partly circular, because the mock validator and the graph share the same metadata.
Future work should test against an independent backend, go beyond SAP reference processes, automate the process layer, and improve seed coverage and the agent's use of tools.


\bibliography{references}

\clearpage
\onecolumn
\setcounter{section}{0}
\setcounter{table}{0}
\setcounter{figure}{0}
\renewcommand{\thesection}{S\arabic{section}}
\renewcommand{\thetable}{S\arabic{table}}
\begin{center}
\large\textbf{Supplementary Information}\\[1mm]
\normalsize \large Grounding LLM Agents in Business Processes: Process-Aware Knowledge Graphs for Reliable Multi-Step Service Invocation
\end{center}
\vspace{2mm}

\noindent These supplementary sections give the secondary analyses and details referred to in the main paper. All $p$ values come from the paired two-sided Wilcoxon signed-rank test over per-query scores, as in the main paper.

\section{Tool Surface Per Condition}
\label{sec:s-tools}

Table~\ref{tab:toolsurface} compares the tools of the two agent conditions. Both use the same mock GET/POST layer and validator in RQ3 and RQ4. They differ only in the retrieval tool and what it returns. Neither condition has a separate schema or value-help tool. Our grounding context arrives inside the retrieval result, while ChunkRAG's retrieval returns summaries only, so ChunkRAG cannot see value-help lists by design. The symmetric baseline in RQ4 (ChunkRAG+VH) adds the value-help lists as plain text to ChunkRAG's retrieval result and changes nothing else.

We do not report a condition with raw tool use and no retrieval. Streaming the \texttt{\$metadata} XML straight into a tool-use prompt timed out in all runs, which would mix up format problems with system performance.

\begin{table}[h]
\centering
\caption{Tool surface per condition, identical across RQ3 and RQ4.}
\label{tab:toolsurface}
\renewcommand{\arraystretch}{1.15}
\begin{tabular}{p{2.6cm}p{5.6cm}p{6.0cm}}
    \toprule
     & \textbf{ChunkRAG} & \textbf{Ours} \\
    \midrule
    Retrieval tool & Dense top-$k$ search over EntitySet summaries~\cite{Pesl:IEEE-TSC:2026} & Dense seeds $+$ bounded process-chain expansion \\
    Retrieval returns & EntitySet, service, and property names & Full typed specification: required keys, navigation cardinality, value-help code lists, process-chain information \\
    Execution & \multicolumn{2}{c}{GET and POST through the same mock HTTP layer and validator} \\
    \bottomrule
\end{tabular}
\end{table}

\section{Benchmark Construction and Annotation Protocol}
\label{sec:s-annotation}

The first version of the benchmark had $80$ queries labelled by one annotator. We grew it to its current size with a two-annotator protocol. First, we generated a pool of $251$ candidate queries automatically from the catalogue and the process graph. Each candidate came with machine-suggested labels and a note on how it was derived. Two annotators then reviewed the full pool independently, following a written protocol. Annotator~A was a human author. Annotator~B was a large language model (Claude, Anthropic); we disclose this openly and release its scripted per-candidate decisions with the repository.

Annotator~A accepted $94$ candidates, revised $156$, and rejected $1$. Annotator~B accepted $100$, revised $149$, and rejected $2$. The annotators agreed on whether to keep or reject $250$ of the $251$ candidates ($99.6\%$), which gives Cohen's $\kappa = 0.665$. The $\kappa$ is only moderate despite near-perfect agreement because almost all candidates were kept. On the $249$ candidates both kept, label agreement over the (query, EntitySet) pairs was $\kappa = 1.000$ ($n = 415$ pairs). Both annotators saw the machine-suggested labels, so this number measures agreement on corrections, not independent labelling.

We discussed the one keep/reject disagreement and dropped the query as unanswerable (it used an opaque internal entity-set name), so $2$ candidates were dropped in total. Where the two accepted wordings differed, we kept the more natural one. We computed agreement before this resolution step and did not recompute it afterwards. The merge added $208$ queries to the main benchmark and $41$ to the elicitation probe. We rebuilt the process-chain ground truth from the graph instead of copying it from the drafts.

The benchmark was split deterministically into $150$ training and $138$ test queries. Some secondary agent analyses (Sec.~\ref{sec:s-agent}) use an earlier 26-query multi-step subset, created before the benchmark grew.

\section{Graph-Construction Fidelity Audit}
\label{sec:s-fidelity}

An independent checker parses the source OData \texttt{\$metadata} and BPMN 2.0 files again and compares the result with the graph. We check each structural class: EntitySets, EntityTypes, Properties, Navigation edges (with cardinality), ValueHelp code lists, Activities, and \texttt{HAS\_NEXT} order. For each class we report coverage and exactness. Coverage is the share of source elements that appear in the graph. Exactness is the share of graph elements whose attributes match the source.

The checker was written separately from the ingestion pipeline and shares none of its parsing logic. It reads OData XML with namespace-qualified queries, while the pipeline matches local element names. It also has its own readers for the SAP catalogue exports and for BPMN.

Two scope notes matter. First, the SAP catalogue is available only as structured CSV exports from the SAP Business Accelerator Hub, not as raw XML, so the audit treats these exports as the source of record. For TripPin, Northwind, and Microsoft Graph, it checks the full path from XML to graph. Second, we compare ValueHelp lists with their hand-curated seed tables, because the ingested \texttt{\$metadata} has no machine-readable value-list annotations. The mock validator uses the same seed tables. The audit does not cover the hand-written \texttt{USES\_ENTITY\_SET} edges, because no independent source exists for them. The main paper assesses them with a provenance split (RQ1).

\begin{table}[h]
\centering
\caption{Graph-construction fidelity audit. Coverage: share of distinct source elements present in the graph. Exactness: share of graph elements whose attributes all match the source. The $79$ activities are the $51$ task activities plus $28$ start and end events.}
\label{tab:fidelity}
\begin{tabular}{lrrrr}
\toprule
\textbf{Structural class} & \textbf{Source} & \textbf{Graph} & \textbf{Coverage} & \textbf{Exactness} \\
\midrule
EntitySets                  &   132 &   132 & 1.000 & 1.000 \\
EntityTypes                 & 1,270 & 1,270 & 1.000 & 1.000 \\
Properties                  & 8,481 & 8,481 & 1.000 & 1.000 \\
Navigation edges (incl.\ cardinality) & 1,499 & 1,499 & 1.000 & 1.000 \\
Activities (BPMN)           &    79 &    79 & 1.000 & 1.000 \\
\texttt{HAS\_NEXT} (BPMN)   &    65 &    65 & 1.000 & 1.000 \\
ValueHelp codes             &    26 &    26 & 1.000 & 1.000 \\
\bottomrule
\end{tabular}
\end{table}

Table~\ref{tab:fidelity} shows the result. Every structural class reaches full coverage and exactness. The audit found only three harmless kinds of difference, and each one comes from the source, not from ingestion. (i)~For $70$ of $8{,}481$ properties, attribute strings differ only by a number or boolean format change from the table-loading step (key flag \texttt{true}\,$\to$\,\texttt{True}; \texttt{MaxLength} \texttt{26}\,$\to$\,\texttt{26.0}). We count these as matches after normalization; they do not change meaning. (ii)~The source tables have $29$ duplicate rows. The loader merges each duplicate into one graph element, and the counts above are over distinct elements. The only attribute-level case is one SAP EntitySet listed twice with two spellings of its entity type; the graph keeps both type links, so nothing is lost. (iii)~\texttt{ComplexType} definitions ($1{,}300$ in Microsoft Graph, $4$ in TripPin) are out of scope by design. The entity model covers only types exposed through EntitySets and the types reachable from them by navigation.

\section{Additional Retrieval Analyses (RQ1)}
\label{sec:s-rq1}

\subsection{Where the gain comes from}

To explain the Hybrid result, we label each gold entity in one of four ways: found by the dense seed, added by process expansion, pushed out by expansion, or missed. Table~\ref{tab:rq1-attr} gives these counts by query type.

\begin{table}[h]
\centering
\caption{Gold-entity attribution for Hybrid at $k=3$ (counts over the 304 gold entities of the test split). \emph{Seed}: in the dense top-3. \emph{Exp.}: only in the Hybrid top-3, added or moved up by process expansion. \emph{Displ.}: in the dense top-3 but pushed out by expansion. \emph{Miss}: in neither. Seed$+$Exp.\ equals Hybrid Recall@3 and Seed$+$Displ.\ equals Dense Recall@3, micro-averaged over gold entities.}
\label{tab:rq1-attr}
\begin{tabular}{lrrrrr}
\toprule
Query type & Golds & Seed & Exp. & Displ. & Miss \\
\midrule
Single entity          &  35 &  33 &  0 &  1 &  1 \\
Value-help dependent   &   1 &   0 &  0 &  0 &  1 \\
Navigation chain       &  49 &  44 &  0 &  1 &  4 \\
Multi-step workflow    &  13 &   6 &  4 &  2 &  1 \\
Process-chain goal     & 206 & 120 & 28 & 12 & 46 \\
\midrule
All                     & 304 & 203 & 32 & 16 & 53 \\
\bottomrule
\end{tabular}
\end{table}

At $k=3$, expansion contributes 32 of the 304 gold entities: 28 in the process-chain probe and four in multi-step workflows. It adds nothing in query types without a process signal. Expansion also pushes 16 dense golds out of the top three. At $k=10$, the dense seed alone finds 271 golds, and the net gain from expansion falls to nine, with two pushed out. This matches the smaller recall gap at larger $k$ in the main RQ1 table.

\subsection{Completeness of the context passed to the agent}

We also check whether each retrieved entity comes with all the context the agent needs. On the agent's discovery path (top-5 seeds), the retrieved context is complete for 3/3 required payload keys, 1/1 value-help code lists, 75/85 navigation cardinalities on gold parent--child pairs, and 67/79 process dependencies. Almost all gaps come from retrieval. In 34 of the 35 failing pairs, at least one entity is missing from the discovery result. Either no dense seed lands in the gold cluster (the \texttt{SalesOrder} cluster in six queries, the Microsoft Graph \texttt{users} cluster in two), or the discovery tool reports process dependencies only for pairs next to a retrieved seed, so second-hop chain members are left out. In the single exception, both entities are retrieved, but their dependency is reported only relative to the dense seeds. Whenever both entities of a required pair are retrieved, keys and cardinality are complete. The gaps are retrieval failures, not failures in building the context.

\subsection{Embedding sensitivity}

With the smaller \texttt{all-mpnet-base-v2} embedding instead of the 8B model, Hybrid still beats Dense on the probe at every $k$ (Recall@3 $0.515$ vs.\ $0.507$; Recall@10 $+3.3$ points). The differences are smaller, however, and none is significant (all $p \geq 0.19$; the paired probe test used in the main paper gives $p=0.54$). Weaker seeds leave more queries out of reach of the graph expansion. The direction of the effect holds, but absolute scores are lower.

\section{Additional Agent Analyses (RQ3)}
\label{sec:s-agent}

The main paper reports RQ3 on the $81$ multi-step test queries. Before the benchmark grew, we ran RQ3 on an earlier 26-query multi-step subset. Several secondary analyses (seed verification, oracle retrieval, and the sensitivity LLMs) were run only on this subset. We report them here.

\begin{table}[h]
\centering
\caption{Agent results on the earlier 26-query subset. ``Invalid/run'' is the mean number of invalid calls per run (\texttt{GET \$metadata} requests are not counted). \texttt{gemma4:31b} runs once per query; the other LLMs run three times. The \texttt{qwen3.6} rows were collected on CPU-only serving with an earlier version of the system and mainly measure serving instability.}
\label{tab:rq3-subset}
\begin{tabular}{llccccc}
    \toprule
    \textbf{LLM} & \textbf{Cond.} & \textbf{Runs} & \textbf{Compl.} & \textbf{Calls/run} & \textbf{Invalid/run} & \textbf{Timeouts} \\
    \midrule
    \multicolumn{7}{l}{\emph{Primary LLM; difference $-0.103$, $p=0.461$ ($n=26$)}} \\
    qwen3-coder:30b & ChunkRAG & 78 & 0.295 & 2.12 & 0.128 & 0/78 \\
    qwen3-coder:30b & Ours     & 78 & 0.192 & 1.69 & 0.051 & 0/78 \\
    \midrule
    \multicolumn{7}{l}{\emph{Sensitivity LLMs}} \\
    gemma4:31b      & ChunkRAG & 26 & 0.192 & 1.00 & 0.000 & 0/26 \\
    gemma4:31b      & Ours     & 26 & 0.115 & 0.65 & 0.000 & 0/26 \\
    qwen3.6 (36B)   & ChunkRAG & 78 & 0.064 & 0.15 & 0.090 & 72/78 \\
    qwen3.6 (36B)   & Ours     & 78 & 0.064 & 0.49 & 0.038 & 54/78 \\
    \bottomrule
\end{tabular}
\end{table}

\subsection{Results on the subset and why they differ}

Table~\ref{tab:rq3-subset} shows the subset results. With the primary LLM, ChunkRAG completed more tasks than our agent ($0.295$ vs.\ $0.192$), but the difference was not significant ($p=0.46$). Our method made fewer invalid calls ($4/78$ vs.\ $10/78$). Two of our four invalid calls happen in one run where in-loop discovery missed the gold entity cluster and the agent guessed from its prior knowledge. The other two happen despite correct retrieval. One invents a property name on a correctly found entity set (\texttt{PurReqnItemText} on \texttt{PurchaseReqnItem}), and one posts \texttt{SalesOrder} under the Northwind service.

On the subset, a wrong seed was the main risk. The offline retriever in RQ1 gets the benchmark query word for word, while the agent retrieves with its own rewording. If the rewording picks a wrong but similar-sounding EntitySet, process expansion adds the wrong chain, and the retrieve-then-execute design fails early. ChunkRAG shows flat endpoint text and may still reach a completing call. Process context therefore helps when the seed is right and makes things worse when it is wrong. On the subset, retrieval misses caused most failures for both methods: $40$ of $55$ ChunkRAG failures and $43$ of $63$ of ours. On the larger test set, which is mostly process-chain goals, ChunkRAG far more often makes no call at all, and this outweighs the wrong-seed risk.

\subsection{Checking seeds before expansion}

We tried the direct fix for committing to a wrong seed too early. Inside the discovery tool, the same LLM checks the dense seeds against the user query, and only verified seeds start process-chain expansion. Unverified seeds are still shown as plain specifications. On the subset ($78$ runs), verification removes invalid writes completely ($0$ invalid POST calls, vs.\ $3$ for the unverified agent and $9$ for ChunkRAG). The cost is more invalid exploratory reads ($11$ GET probes to service/entity pairs that do not exist, which the mock rejects harmlessly). Completion improves only slightly ($0.205$ vs.\ $0.192$; ChunkRAG $0.295$). The conditional rates explain why. Completion is $0.421$ when in-loop discovery returned all gold entities and $0.000$ when it did not. Verification can reject a wrong seed but cannot add a missing one, and the main failure is a \emph{missing} seed, not a wrong one. Closing the completion gap therefore needs better seed coverage (larger candidate pools, query rewriting), not stricter filtering.

\subsection{Separating execution from retrieval (oracle retrieval)}

RQ3 mixes retrieval and execution failures. We therefore add an oracle condition in which the agent receives the gold specification for each query (perfect retrieval). Any remaining failure must then come from payload construction and tool use. We score required-key adherence, value-help adherence, correct nesting of 1:N navigations as arrays, and retries after the validator rejects a call, for both conditions under the same oracle context.

Under oracle retrieval ($78$ runs per condition, no timeouts), completion stays low for both agents (ours $0.256$, ChunkRAG $0.359$). The completion gap on the subset is thus not only a retrieval effect. It also reflects execution limits of the 30B LLM, mainly failures to use the tools and incomplete payloads. Payload construction still separates the two conditions. With the same gold context, ChunkRAG issues $15$ schema-invalid writes, including $6$ invented codes, while ours issues $3$ and invents none. Required-key adherence is low for both ($0.25$ vs.\ $0.08$). Only the typed prompt leads to correct nesting of 1:N children as arrays ($6$ of $29$ posting runs vs.\ $0$ of $22$). Retrying after a rejected call is rare in both conditions ($2$ vs.\ $0$ runs). Looking only at runs where retrieval succeeded gives the same picture. Over the $35$ runs whose discovery returned every gold entity, completion is $0.43$ (ours) vs.\ $0.57$ (ChunkRAG), against $0.00$ vs.\ $0.07$ when it did not. Good retrieval is necessary for completion, but execution remains the main bottleneck.

\subsection{Sensitivity LLMs and serving}

We use the non-primary LLMs only to check external validity. \texttt{gemma4:31b} runs once per query ($26$ runs per condition, no timeouts). On the subset, its results point the same way as the primary LLM: higher ChunkRAG completion, and zero invalid calls for both. \texttt{qwen3.6} hit the latency ceiling in $92\%$ of ChunkRAG runs and $69\%$ of our runs, so its rows mainly measure serving instability rather than agent behaviour. Local LLM serving can stall across condition switches when its cache is reused on CPU-only serving; restarting the server between conditions avoids this. On GPU serving (final runs), no restarts were needed. Our conclusions rest on the primary \texttt{qwen3-coder:30b}, which had $0$ timeouts in $642$ runs.

\section{Mapping Authoring}
\label{sec:s-mapping}

\subsection{Authoring procedure}

Authoring followed a fixed procedure. For each BPMN task activity, the author (1) identifies the business documents the activity reads or writes, (2) finds the covering service in the catalogue, (3) checks the exact EntitySet names against that service's \texttt{\$metadata}, and (4) records the (activity, service, EntitySet) triples in a version-controlled mapping file. Activities with no covering service are left unmapped and documented in the file.

For the timing study, one author familiar with the catalogue re-wrote a random sample (fixed seed) of ten mapped activities, covering $12$ of the $45$ edges, from scratch under a timer, using this procedure. The sample took $205$ person-minutes (median $20.5$ minutes per activity, range $16$--$25$), and every re-written edge matched the released mapping.

\subsection{LLM-proposed mapping}

For each of the $14$ processes, a local LLM (\texttt{qwen3-coder:30b}) receives the catalogue (services with their entity sets) and the ordered task activities, and proposes \texttt{USES\_ENTITY\_SET} edges. Against the $45$ hand-written edges, the proposals reach precision $0.79$ and recall $0.93$ (F1 $0.86$). They recover $42$ of the $45$ gold edges. The $11$ false positives are plausible mappings that a reviewer can reject quickly, because no service in the catalogue covers them (e.g., mapping Hire-to-Retire tasks to the Microsoft Graph \texttt{users} entity, or goods receipt to \texttt{PurchaseOrderItem}). Authoring the mapping thus becomes checking $53$ proposed edges instead of writing $45$ from scratch. The three misses (the TripPin flight-lookup entities and one quality-notification entity) show that domain review is still needed.

\section{Per-Sector Results on SOCBench-D (RQ5)}
\label{sec:s-sectors}

\begin{table}[h]
\centering
\caption{RQ5 per-sector Recall@5 on SOCBench-D (Hybrid vs.\ Tokens), $10$ queries per GICS sector.}
\label{tab:rq4sectors}
\begin{tabular}{lrrr}
    \toprule
    \textbf{Sector} & \textbf{Hybrid} & \textbf{Tokens} & $\Delta$ \\
    \midrule
    01 Energy                  & 0.346 & 0.258 & $+0.088$ \\
    02 Materials               & 0.533 & 0.432 & $+0.101$ \\
    03 Industrials             & 0.364 & 0.307 & $+0.058$ \\
    04 Consumer discretionary  & 0.502 & 0.477 & $+0.025$ \\
    05 Consumer staples        & 0.397 & 0.394 & $+0.003$ \\
    06 Health care             & 0.339 & 0.403 & $-0.064$ \\
    07 Financials              & 0.471 & 0.517 & $-0.046$ \\
    08 Information technology  & 0.384 & 0.505 & $-0.121$ \\
    09 Communication services  & 0.346 & 0.531 & $-0.184$ \\
    10 Utilities               & 0.310 & 0.399 & $-0.089$ \\
    11 Real estate             & 0.320 & 0.323 & $-0.003$ \\
    \bottomrule
\end{tabular}
\end{table}

The per-sector results are mixed. Hybrid is ahead in five sectors and behind in six. The differences depend on how closely endpoint paths match the words in the queries. Without a process layer, the graph adds no consistent benefit, as the main paper predicts.

\end{document}